\documentclass[11pt]{article}

\usepackage[preprint]{acl}

\usepackage{times}
\usepackage{latexsym}

\usepackage[T1]{fontenc}

\usepackage[utf8]{inputenc}

\usepackage{microtype}

\usepackage{inconsolata}

\usepackage{graphicx}

\usepackage{booktabs}
\usepackage{amsmath}
\usepackage{subcaption}
\usepackage{siunitx}
\usepackage{enumitem}
\usepackage{placeins}

\usepackage{tcolorbox}
\tcbuselibrary{listings}
\newtcolorbox{promptbox}[1][]{
  colback=gray!5,
  colframe=gray!50,
  fonttitle=\bfseries,
  coltitle=black,
  title=#1,
  boxrule=0.5pt,
  arc=2mm,
  outer arc=2mm,
  left=4pt,right=4pt,top=4pt,bottom=4pt,
  after skip=12pt,
  listing only,
  listing options={
    basicstyle=\ttfamily\small,
    breaklines=true,
    columns=fullflexible
  }
}

\title{Dating the Model: Hidden Dates in System Prompts\\Affect LLM Evaluation}

\author{
    Mario Sanz-Guerrero$^{1}$\quad
    Minh Duc Bui$^{1}$\quad
    \bf Manuel Mager$^{2}$\quad
    Katharina von der Wense$^{1,3}$\\
    $^1$Johannes Gutenberg University Mainz, Germany\\
    $^2$Universidad Iberoamericana, Mexico\\
    $^3$University of Colorado Boulder, USA\\
    \texttt{\href{mailto:msanz@uni-mainz.de}{msanz@uni-mainz.de}}
}

\begin{document}
\maketitle
\begin{abstract}
Reproducibility is essential for scientific research, yet prior work shows that LLM outputs vary with hardware and batching. We identify an overlooked factor:
the hidden injection of the current date into system prompts, which users cannot control and which changes every day.
Across 9 recent LLMs and 6 datasets spanning multiple-choice QA (MCQA), math reasoning, code generation, and machine translation, performance varies \emph{solely} with the current date, with deltas of up to 6\% on MCQA, 14\% on math reasoning, 7\% on code generation, and 2.84 BLEU on machine translation. Model rankings also shift, affecting leaderboards.
This date effect exceeds other sources of non-determinism, such as batch size and numerical precision. Standard prompting techniques -- chain-of-thought and few-shot prompting -- do not reduce the sensitivity; chain-of-thought even amplifies it.
Our findings underscore the need for careful evaluation protocols to ensure reproducibility and fair comparisons in LLM research.
\end{abstract}

\section{Introduction}

\begin{figure}[t]
    \centering
    \begin{subfigure}{\columnwidth}
        \centering
        \includegraphics[width=\textwidth]{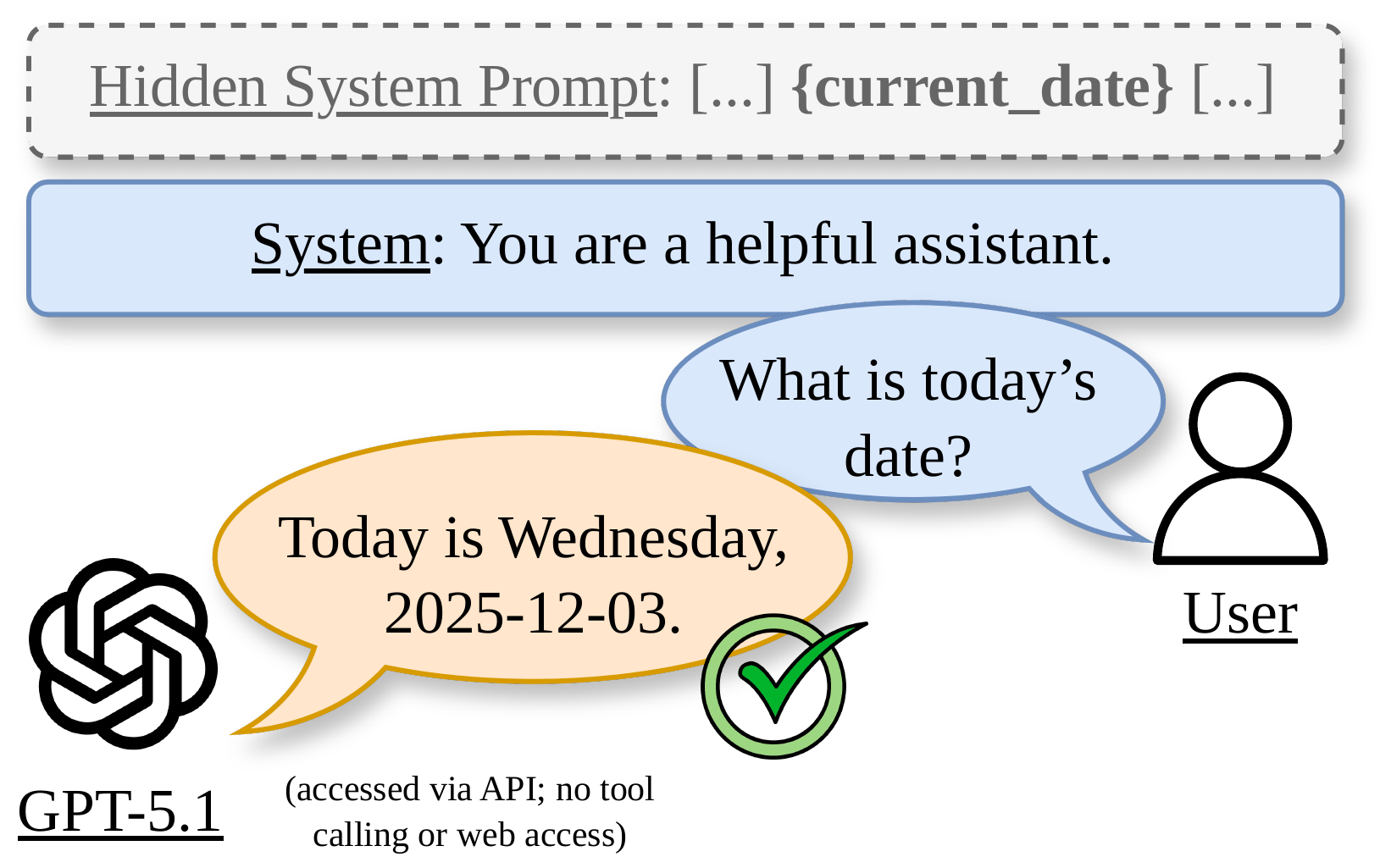}
        \caption{GPT-5.1 knows today's date, even though the date is not included in the user-controllable prompt (represented in blue).}
        \label{subfig:gpt5_date}
    \end{subfigure}

    \begin{subfigure}{\columnwidth}
        \centering
        \includegraphics[width=\textwidth]{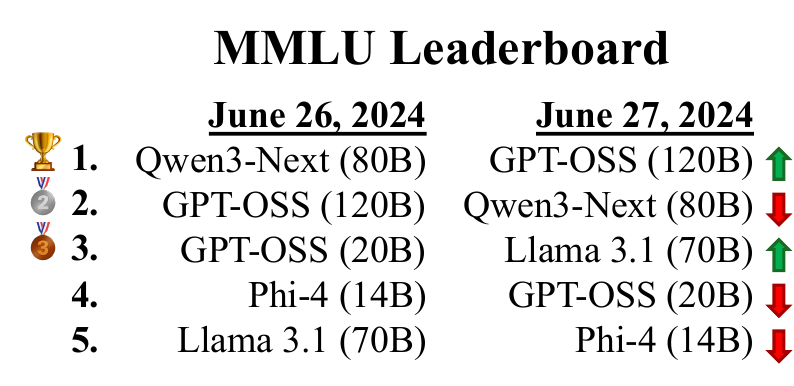}
        \caption{Top-5 models on MMLU on two consecutive dates: changing only the date reorders the leaderboard.}
        \label{subfig:leaderboard_date}
    \end{subfigure}

    \caption{The performance of LLMs on standard benchmarks varies solely by changing the current date, which is usually hidden from users in the system prompt.}
    \label{fig:date_problem}
\end{figure}

The rapid progress of large language models (LLMs) requires careful benchmarking, but reproducibility -- a crucial aspect of scientific research -- is hindered by the non-determinism\footnote{We adopt this term to align with the literature, though technically speaking, determinism is achievable under identical configurations. See Appendix \ref{app:non_determinism} for a detailed discussion.} of LLM outputs. Prior work attributes this to factors such as inference batch size, numerical precision, and hardware \cite{he2025nondeterminism,yuan2025understanding}.

In this paper, we identify and analyze a so-far overlooked source of non-determinism: the hidden injection of the current date into the system prompt. Since this information is time-varying, a critical question arises: \emph{do evaluations fluctuate on different days, even with identical settings?} Figure~\ref{fig:date_problem} illustrates this phenomenon.

Across 9 recent LLMs and 6 standard benchmarks spanning four tasks, we observe accuracy variations up to 6\% on multiple-choice QA (MCQA) \emph{just} from changing the date, and model rankings reorder accordingly (Figure \ref{subfig:leaderboard_date}). The effect is even larger on language-generation tasks, reaching 14\% on math reasoning, 7\% on code generation, and 2.84 BLEU on machine translation. This date effect is larger than other known sources of non-determinism, such as batch size and numerical precision. Common prompting techniques do not remove it either: neither chain-of-thought nor few-shot prompting reduces the sensitivity, and chain-of-thought even makes it worse.
Our findings underscore the need for evaluation protocols that account for hidden prompt metadata to ensure reproducibility and fair comparisons in LLM research.

\section{Related Work}
There is a growing body of work studying non-determinism in LLM evaluations and its implications for reproducibility.
Recent studies highlight that LLM benchmarks are highly sensitive to configuration choices. \citet{song2025goodbadgreedy} and \citet{atil2025nondeterminismdeterministicllmsettings} report that even ``deterministic'' greedy decoding yields unstable results, while \citet{hochlehnert2025reproducibility} find that reasoning performance fluctuates widely based on subtle implementation details, including random seeds, prompt structure, and decoding hyperparameters like temperature.
\citet{sclar2024quantifying} and \citet{sanzguerrero2025mindthegap} show that minor changes to prompt formatting (e.g., separators or spacing) substantially affect model performance.
At the system level, \citet{yuan2025understanding} and \citet{he2025nondeterminism} show that inference batch size, numerical precision, and hardware differences introduce variability in LLM outputs due to floating-point arithmetic errors.
Prior work largely attributes non-determinism to batching, hardware, or prompt formatting; we identify a distinct source of variance: the hidden, time-varying insertion of the current date into system prompts.

\section{Experimental Setup}

\paragraph{Prompts}
To isolate the effect of the current date -- a detail often hidden from users -- we use identical prompts for all models, changing \emph{only} the date in the system prompt and keeping the rest of the configuration fixed and deterministic (see Appendix~\ref{app:experimental_details} for details). We sweep dates from January 1 to December 31, 2024, covering a representative year during which LLMs were rapidly developed, improved, and benchmarked.

\paragraph{Datasets}
We use standard datasets commonly reported in new model releases. For multiple-choice QA, we experiment on MMLU \cite{hendrycks2021mmlu}, GPQA \cite{rein2024gpqa}, and ARC-Challenge \cite{clark2018arc}. We verify that none of the questions are time-dependent to avoid confounding factors (see Appendix \ref{app:not_time_dependent}).
In MCQA, the prediction comes from the probability of a single answer token, which gives the date only a small surface to act on. To test whether the effect grows when the model generates longer outputs, we also evaluate on GSM8K \citep{cobbe2021gsm8k} math reasoning, where the model produces a step-by-step solution and we grade the final number.
However, GSM8K still reduces evaluation to a single extracted number, so we further test code generation on HumanEval \citep{chen2021humaneval}, where the model writes a complete Python function that we run against unit tests (i.e., correctness depends on the whole generated program).
Finally, to test the effect when the \emph{full} output is evaluated, we run all models on machine translation (MT) with three pairs from WMT \citep[English to German, Finnish, and Czech;][]{wmt16}.

\paragraph{Models}
We evaluate 9 recent LLMs from various families, sizes and capabilities:
Llama 3.1 Instruct \citep[8B \& 70B;][]{grattafiori2024llama3}, Gemma 3 Instruct \citep[4B \& 27B;][]{gemma2025gemma3}, Qwen3 \citep[4B;][]{yang2025qwen3}, Qwen3-Next \citep[80B;][]{qwen2025qwen3next}, Phi-4 \citep[14B;][]{abdin2024phi4}, and GPT-OSS \citep[20B \& 120B;][]{openai2025gptoss}.

\paragraph{Evaluation}
For MCQA, we report accuracy and calibration. Calibration is measured via expected calibration error \citep[ECE;][]{naeini2015ece}, the weighted absolute gap between accuracy and confidence across $M=10$ equal-width bins (see Appendix \ref{app:ece} for the formula). For GSM8K, we report accuracy; for HumanEval, we report pass@1; and, for MT, we use BLEU \citep{papineni2002bleu} and chrF \citep{popovic2015chrf}.

\section{Results}

\begin{figure*}[h]
    \centering
    \includegraphics[width=\textwidth]{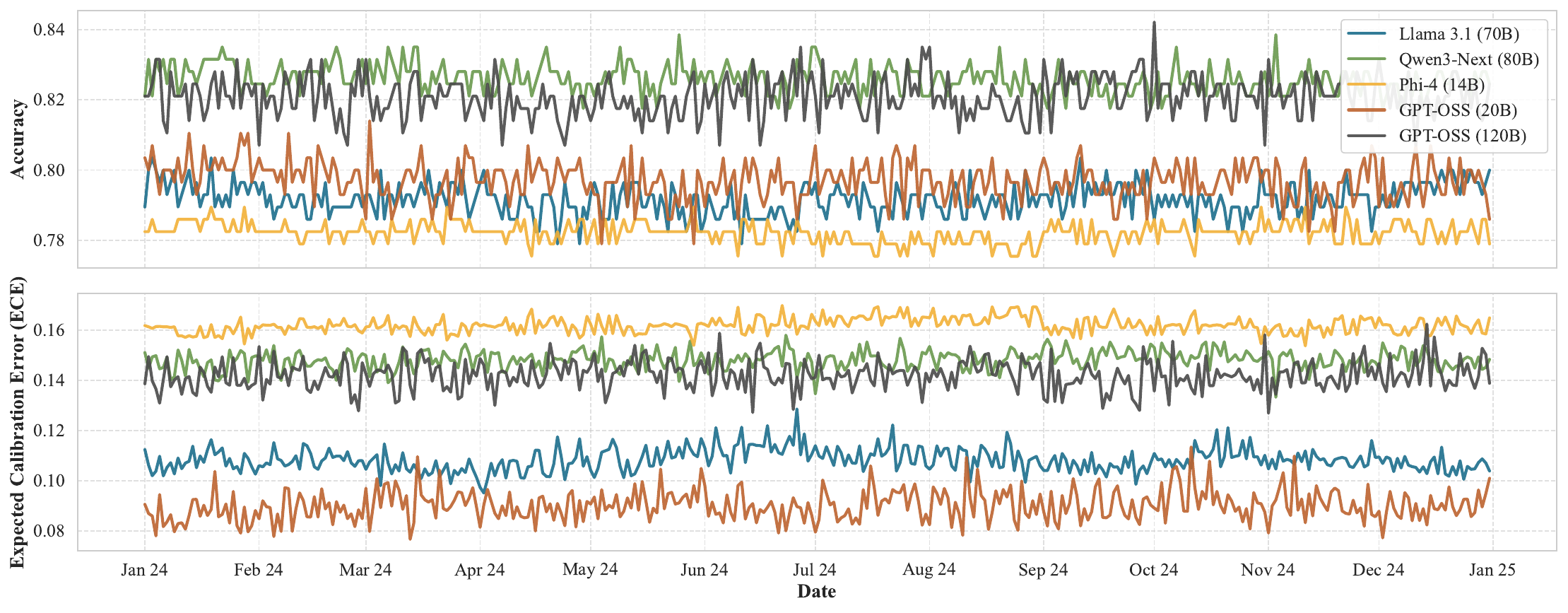}
    \caption{Accuracy (top) and ECE (bottom) across different dates in 2024 for the top-5 models on MMLU.}
    \label{fig:accuracy_ece_top5}
\end{figure*}

\paragraph{MCQA Results Vary with Date Changes}
Figure \ref{fig:accuracy_ece_top5} summarizes accuracy and ECE across dates in 2024 for the top-5 models on MMLU (full results in Appendix \ref{app:detailed_results}). Although the current date should be irrelevant, both accuracy and calibration vary substantially with it, and model rankings reorder (see Figure \ref{subfig:leaderboard_date}).
Table \ref{tab:delta_accuracy_dates} reports the worst-to-best accuracy delta across models and datasets. Differences reach up to 6\%, which is substantial given that the \emph{only} changing factor is the current date in the system prompt and all questions are time-independent (see Appendix \ref{app:not_time_dependent}).

\begin{table}
    \centering
    \small
    \setlength{\tabcolsep}{4pt}
    \begin{tabular}{l S[table-format=1.2] S[table-format=1.2] S[table-format=1.2] | S[table-format=1.2]}
        \toprule
        Model & {MMLU} & {GPQA} & {ARC-C} & {Avg.}\\
        \midrule
        Llama 3.1 (8B) & 2.11 & 4.53 & 2.01 & 2.88 \\
        Llama 3.1 (70B) & 2.46 & 3.54 & 0.67 & 2.22 \\
        Gemma 3 (4B) & 1.75 & 3.03 & 1.87 & 2.22 \\
        Gemma 3 (27B) & 1.38 & 4.04 & 1.01 & 2.14 \\
        Qwen3 (4B) & 2.46 & 2.53 & 1.34 & 2.11 \\
        Qwen3-Next (80B) & 2.11 & 2.43 & 1.01 & 1.85 \\
        Phi-4 (14B) & 1.40 & 3.51 & 0.33 & 1.75 \\
        GPT-OSS (20B) & 3.49 & 4.55 & 2.34 & 3.46 \\
        GPT-OSS (120B) & 3.51 & 6.06 & 2.53 & 4.03 \\
        \midrule
        Average & 2.30 & 3.80 & 1.46 & 2.52 \\
        \bottomrule
    \end{tabular}
    \caption{Difference in accuracy (delta) from the worst to the best date in 2024 across models and datasets.}
    \label{tab:delta_accuracy_dates}
\end{table}

\paragraph{The Effect Is Larger for Reasoning Tasks}
\label{subsec:gsm8k}
The GSM8K column of Table \ref{tab:gen_results} shows deltas larger than on MCQA -- 7.75\% on average vs.\ 2.52\% in Table \ref{tab:delta_accuracy_dates}. Even the largest models are affected, so scale does not protect against the effect. This matches the mechanism we investigate further in Section \ref{sec:cot_amplifies}: longer autoregressive generations give the date prefix repeated opportunities to bias intermediate tokens, and these perturbations cascade into different final answers. Since reasoning benchmarks like GSM8K are central to current leaderboards, deltas of this magnitude make accuracy reported on different days difficult to compare.

\begin{table*}
    \centering
    \small
    \begin{tabular}{l S[table-format=2.2] | S[table-format=1.2] | S[table-format=1.2] S[table-format=1.2] S[table-format=1.2] S[table-format=1.2] S[table-format=1.2] S[table-format=1.2]}
        \toprule
        & {GSM8K} & {HumanEval} & \multicolumn{2}{c}{WMT (en $\to$ de)} & \multicolumn{2}{c}{WMT (en $\to$ fi)} & \multicolumn{2}{c}{WMT (en $\to$ cs)} \\
        \cmidrule(lr){2-2} \cmidrule(lr){3-3} \cmidrule(lr){4-5} \cmidrule(lr){6-7} \cmidrule(lr){8-9}
        Model & {$\Delta$ Acc.} & {$\Delta$ pass@1} & {$\Delta$ BLEU} & {$\Delta$ chrF} & {$\Delta$ BLEU} & {$\Delta$ chrF} & {$\Delta$ BLEU} & {$\Delta$ chrF} \\
        \midrule
        Llama 3.1 (8B)   &  9.85 & 7.32 & 1.66 & 1.13 & 1.79 & 1.52 & 1.82 & 1.42 \\
        Llama 3.1 (70B)  &  7.58 & 5.49 & 1.58 & 1.07 & 1.71 & 1.46 & 1.76 & 1.36 \\
        Gemma 3 (4B)     &  8.33 & 6.71 & 1.57 & 1.15 & 1.64 & 1.11 & 1.68 & 1.22 \\
        Gemma 3 (27B)    &  3.03 & 3.66 & 1.58 & 0.96 & 1.38 & 0.91 & 1.90 & 1.06 \\
        Qwen3 (4B)       &  6.06 & 2.44 & 1.34 & 0.95 & 1.39 & 1.38 & 1.88 & 1.30 \\
        Qwen3-Next (80B) &  4.55 & 1.83 & 1.28 & 0.90 & 1.31 & 1.29 & 1.77 & 1.21 \\
        Phi-4 (14B)      &  3.03 & 1.83 & 0.47 & 0.27 & 0.47 & 0.45 & 0.58 & 0.38 \\
        GPT-OSS (20B)    & 12.88 & 7.32 & 2.35 & 1.40 & 1.91 & 1.77 & 2.66 & 1.91 \\
        GPT-OSS (120B)   & 14.42 & 6.71 & 2.52 & 1.56 & 2.14 & 1.93 & 2.84 & 2.08 \\
        \midrule
        Average          &  7.75 & 4.81 & 1.59 & 1.04 & 1.53 & 1.31 & 1.88 & 1.33 \\
        \bottomrule
    \end{tabular}
    \caption{Worst-to-best date delta in 2024 across models on open-ended generation tasks: accuracy on GSM8K, pass@1 on HumanEval, and BLEU and chrF on 3 WMT language pairs (English to German, Finnish, and Czech).}
    \label{tab:gen_results}
\end{table*}

\paragraph{Execution-Graded Code Still Shifts}
\label{subsec:humaneval}
The HumanEval column of Table \ref{tab:gen_results} shows the same pattern for code generation. Pass@1 changes by 4.81\% on average just from the date, and by up to 7.32\% for the most affected models. This is notable because code is graded by running it against unit tests, so the metric is objective and does not depend on the surface form of the output. Even so, the date still moves the results, so the effect reaches a very different task with a strict, execution-based metric.

\paragraph{The Effect Holds for Full-Output Metrics}
\label{subsec:mt}

The right part of Table \ref{tab:gen_results} shows that the effect persists for MT, with average deltas of up to 1.88 BLEU and 1.33 chrF. Since the setup is fully deterministic, this variability is attributable solely to the hidden date. This is not a small effect in MT, where progress is often reported in fractions of a BLEU point. Together with GSM8K and HumanEval, this shows the effect is not an MCQA artifact but a general property of LLM evaluation under hidden, time-varying prompt metadata.

\section{Analysis}

\subsection{No Date Is Consistently Better}
\label{subsec:patterns}
Figure~\ref{fig:accuracy_ece_top5} shows no obvious pattern in which dates help or hurt, so we analyze this systematically on MCQA (Table~\ref{tab:date_patterns}). We test three patterns: (i) a \emph{trend over the year}, via the Spearman correlation between the date and accuracy for each of the 27 model--dataset combinations (9 models $\times$ 3 datasets); (ii) an effect \emph{shared across datasets}, via the Pearson correlation between the accuracies across dates of the same model on two datasets (9 models $\times$ 3 dataset pairs $=$ 27 comparisons); and (iii) an effect \emph{shared across models}, via the Pearson correlation between the accuracies across dates of two models on the same dataset (36 model pairs $\times$ 3 datasets $=$ 108 comparisons). A correlation near zero means no pattern. For (ii) and (iii), we also report how often a date moves both accuracies in the same direction, i.e., both above or both below their yearly average, where 50\% corresponds to chance.

We observe no consistent pattern.
Over time, the correlation between date and accuracy is close to zero for most model--dataset combinations (median 0.02), with no common direction.
Across datasets and across models, correlations are also centered at zero, and a date moves both accuracies in the same direction on only 51\% of dates -- what we expect by chance. This holds even for models of the same family, which share the tokenizer and chat template. Hence, there is no ``good'' date to fix for evaluation: the date acts as noise specific to each model and dataset.

\begin{table}
    \centering
    \small
    \setlength{\tabcolsep}{2.5pt}
    \begin{tabular}{l c c c c}
        \toprule
        Pattern & {N} & {Median} & {Range} & {Same dir.} \\
        \midrule
        Trend over the year & 27 & 0.02 & [$-$0.40, 0.53] & -- \\
        \cmidrule(lr){1-5}
        Across datasets & 27 & $-$0.03 & [$-$0.16, 0.15] & 51\% \\
        Across models & 108 & 0.01 & [$-$0.17, 0.25] & 51\% \\
        \bottomrule
    \end{tabular}
    \caption{Correlation between the date and accuracy (trend over the year; Spearman), and between the accuracies across dates of the same model on two datasets or of two models on the same dataset (Pearson), on MCQA. \emph{N}: number of model--dataset combinations (trend) or compared pairs. \emph{Same dir.}: share of dates where both accuracies are above or both below their yearly average (50\% = chance).}
    \label{tab:date_patterns}
\end{table}

\subsection{The Effect Reaches Proprietary Models}
\label{subsec:gpt5}
\begin{figure}[h]
    \centering
    \includegraphics[width=\columnwidth]{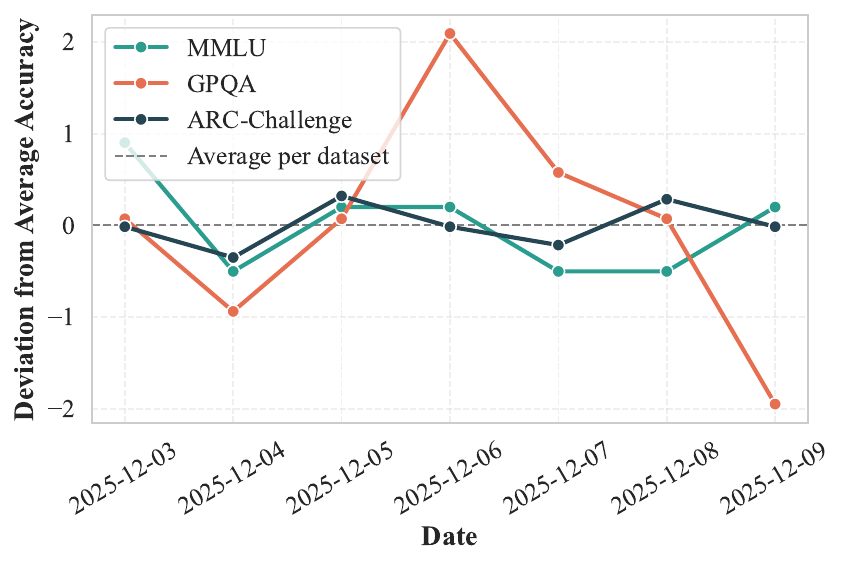}
    \caption{Accuracy deviation across different dates for GPT-5.1. The 0.0 line represents the average accuracy over the week for each dataset.}
    \label{fig:gpt5_accuracy_deviation}
\end{figure}

To validate our findings on a proprietary model, we evaluate GPT-5.1\footnote{Specific checkpoint: \texttt{gpt-5.1-2025-11-13}.} over one week (December 3--9, 2025) on our MCQA datasets. We leave the system prompt empty -- the date is injected server-side (see Figure \ref{subfig:gpt5_date}) -- set the temperature to 0, and disable reasoning.\footnote{We run each date twice and obtain identical scores in all 42 runs (7 days $\times$ 3 datasets $\times$ 2 repetitions), which strongly suggests that the model is deterministic at a fixed date. Therefore, the variation we report comes from the date change.}
Figure \ref{fig:gpt5_accuracy_deviation} shows that GPT-5.1 is also affected, with variations of up to 4\% (on GPQA).

\subsection{Chain-of-Thought Amplifies Date Sensitivity}
\label{sec:cot_amplifies}
MCQA is typically evaluated by reading the answer from the next-token probability \cite{lm-eval-harness}, an efficient but, as we have shown, date-sensitive setup. We further test whether chain-of-thought (CoT) prompting \cite{wei2022cot} mitigates this, as explicit step-by-step reasoning could ground the model and reduce the influence of superficial metadata. We evaluate Llama 3.1 (8B) with CoT on MMLU across all dates in 2024.

\begin{figure}
    \centering
    \includegraphics[width=\columnwidth]{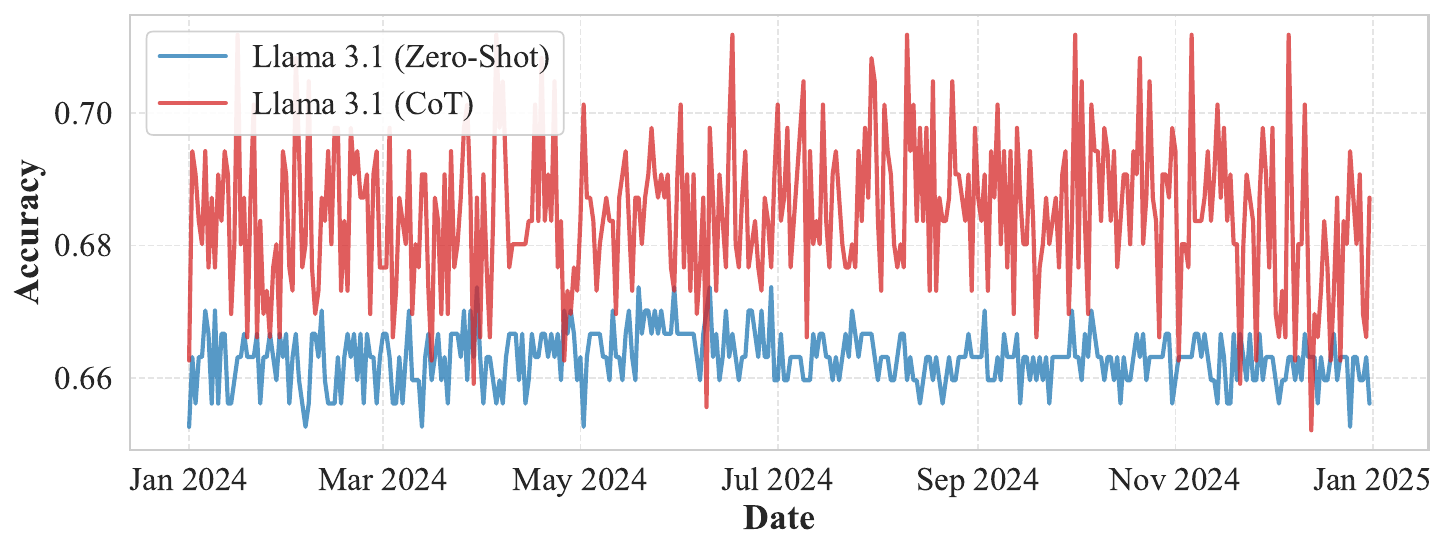}
    \caption{Comparison of accuracy across different dates in 2024 for Llama 3.1 (8B) on MMLU using zero-shot prompting vs.\ chain-of-thought prompting.}
    \label{fig:cot_vs_zero_shot}
\end{figure}

Contrary to our hypothesis, Figure \ref{fig:cot_vs_zero_shot} shows that CoT \emph{amplifies} the date sensitivity. We observe that the date affects the selection of initial CoT tokens, and, due to the autoregressive nature of LLMs, these small perturbations cascade into different reasoning paths and different final answers. So open-ended generation gives the date even more surface to act as a confounder.

\subsection{Few-Shot Learning Does Not Help Either}
Few-shot prompting is another plausible mitigation: task examples could ground the model and reduce the unintentional influence of the date. However, with 5-shot prompting (Table \ref{tab:results_few_shot}), the gaps persist across all models, with an average accuracy delta of 2.27\% (vs.\ 2.52\% zero-shot).

\begin{table}
    \centering
    \small
    \setlength{\tabcolsep}{4pt}
    \begin{tabular}{l S[table-format=1.2] S[table-format=1.2] S[table-format=1.2] | S[table-format=1.2]}
        \toprule
        Model & {MMLU} & {GPQA} & {ARC-C} & {Avg.}\\
        \midrule
        Llama 3.1 (8B) & 2.81 & 5.05 & 2.35 & 3.40 \\
        Llama 3.1 (70B) & 1.40 & 5.56 & 1.01 & 2.66 \\
        Gemma 3 (4B) & 2.46 & 3.54 & 1.34 & 2.45 \\
        Gemma 3 (27B) & 0.70 & 1.52 & 0.67 & 0.96 \\
        Qwen3 (4B) & 1.75 & 3.03 & 1.43 & 2.07 \\
        Qwen3-Next (80B) & 1.75 & 3.54 & 1.34 & 2.21 \\
        Phi-4 (14B) & 1.40 & 2.02 & 0.31 & 1.24 \\
        GPT-OSS (20B) & 2.11 & 2.53 & 0.67 & 1.77 \\
        GPT-OSS (120B) & 3.51 & 4.55 & 3.02 & 3.69 \\
        \midrule
        Average & 1.99 & 3.48 & 1.35 & 2.27 \\
        \bottomrule
    \end{tabular}
    \caption{Difference in accuracy (delta) from the worst to the best date in 2024 across models and datasets using 5-shot prompting.}
    \label{tab:results_few_shot}
    \vspace{-7pt}
\end{table}

\subsection{Comparison Against Other Sources of Non-Determinism}
To assess how the date impact compares in magnitude to other sources of non-determinism in LLM evaluations, we benchmark it against the factors highlighted in prior work \cite{he2025nondeterminism,yuan2025understanding,zheng2024llmnotrobust,pezeshkpour2024optionorder}: inference batch size (1--128 in powers of 2), numerical precision (BF16, FP16, FP32), GPU model (A100, A40, RTX4090), and option order (5 random permutations per question). All runs use Llama 3.1 (8B) on MMLU and an A100 (except for comparing GPUs). We quantify variation via the coefficient of variation (CV; standard deviation over mean) of accuracy and ECE, which normalizes by the mean and puts all sources on a comparable scale.

Table \ref{tab:nondeterminism_comparison} shows that, while batch size, numerical precision, and hardware do introduce variability, their effect is consistently smaller than that of the date. Option order -- a well-known source of non-determinism \cite{zheng2024llmnotrobust,pezeshkpour2024optionorder} -- is comparable in magnitude to the date effect.

\begin{table}
    \centering
    \small
    \setlength{\tabcolsep}{4pt}
    \begin{tabular}{l S[table-format=1.2, table-space-text-post={\%}, detect-weight] S[table-format=1.2, table-space-text-post={\%}, detect-weight]}
        \toprule
        Source of Variation & {Acc.} & {ECE} \\
        \midrule
        Current Date \scriptsize{[Jan 1 -- Dec 31]} & 0.78\% & 2.56\% \\
        \cmidrule(lr){1-3}
        Batch Size \scriptsize{[1, 2, 4, 8, \dots, 128]} & 0.61\% & 2.00\% \\
        Num.\ Precision \scriptsize{[BF16, FP16, FP32]} & 0.52\% & 2.54\% \\
        GPU Model \scriptsize{[A100, A40, RTX4090]} & 0.48\% & 1.60\% \\
        Option Order \scriptsize{[5 random permutations]} & 0.75\% & 3.12\% \\
        \cmidrule(lr){1-3}
        System Instruction \scriptsize{[6 different wordings]} & 0.78\% & 3.54\% \\
        \bottomrule
    \end{tabular}
    \caption{Coefficient of variation (CV) of accuracy and ECE from different sources of non-determinism using Llama 3.1 (8B) on MMLU.}
    \label{tab:nondeterminism_comparison}
\end{table}

\subsection{Other Variations in the System Prompt}
One possible reason for the date sensitivity is that the system prompt might be fixed during supervised fine-tuning (SFT) and reinforcement learning from human feedback (RLHF), making the model brittle to any slight modification. To test this, we vary the system prompt's wording across six versions (see Appendix \ref{app:wordings}).
The last row of Table \ref{tab:nondeterminism_comparison} shows that the resulting variation has the same accuracy CV as the date effect (0.78\%). The hidden, dynamic change of the current date has an impact on model behavior similar to that of intentional prompt engineering.
This supports the hypothesis that the model is sensitive to any system-prompt change, and the date is such a change -- except that it happens without the user's knowledge.

\section{Conclusion}
We identify a critical, often overlooked source of non-determinism in LLM evaluation: the hidden injection of the current date into system prompts.
Across 9 models, 6 datasets, and four tasks, this dynamic metadata alters model performance and reshuffles leaderboard rankings, surpassing the variance introduced by other system-level factors such as batch size or numerical precision. The effect is larger for language-generation tasks: up to 6\% accuracy on MCQA, 14\% accuracy on math reasoning, 7\% pass@1 on code generation, and 2.84 BLEU on machine translation.
Neither CoT nor few-shot prompting reduces this sensitivity; CoT in fact amplifies it.
In practice, we recommend removing the date from the chat template when possible while keeping the rest of the template unchanged, or otherwise fixing the date and reporting it alongside the results.
These findings highlight the fragility of current benchmarking protocols and the necessity of accounting for hidden prompt metadata to ensure reproducibility.

\section*{Limitations}
Our study shows that the injection of the current date into system prompts can significantly affect LLM evaluation results, highlighting a critical source of non-determinism which is dynamic over time. However, our experiments are limited to a select set of models and datasets, and our proprietary-model evaluation is limited to GPT-5.1 over one week. Further research is needed to generalize these findings across a broader range of LLMs and tasks.
To keep the scope (and costs) manageable, we cover four representative evaluation tasks -- multiple-choice QA, math reasoning (GSM8K), code generation (HumanEval), and machine translation -- since our experimental setup requires 365 runs (per model, per dataset) to cover all dates in a year. Other open-ended tasks such as dialogue or summarization remain to be studied. We also inject the date in a fixed format and at a fixed position in the prompt and vary it within 2024; sensitivity to other formats, positions, or years remains to be explored.

\section*{Acknowledgments}
This work was supported by the Carl Zeiss Foundation through the MAINCE and TOPML projects (grant numbers P2022-08-009 and P2021-02-014).

\bibliography{refs}

\clearpage

\appendix

\section{On the Use of the Term ``Non-Determinism''}
\label{app:non_determinism}
In this work, we adopt the term ``non-determinism'' to describe the variability in LLM outputs when determinism is expected, aligning with its widespread usage in recent literature \cite{he2025nondeterminism,yuan2025understanding,song2025goodbadgreedy,hochlehnert2025reproducibility,atil2025nondeterminismdeterministicllmsettings}.

From a strictly technical perspective, however, standard LLMs are deterministic functions. Determinism is achievable provided that the entire configuration remains identical. Indeed, in our own experiments, we observed that when we hold the current date in the system prompt fixed, the model returns identical outputs across multiple runs.

The practical reality for researchers, however, is different from this theoretical determinism. Critical system configurations (e.g., batch size during inference, floating-point numerical precision, or GPU model) are frequently out of the user's control, particularly in API-based environments. Furthermore, these low-level details are rarely reported in experimental setups.
Even if they were extensively documented, differences in hardware or runtime environment can lead to divergent outputs.
As a result, two independent studies aiming to replicate each other's results may not operate under truly identical conditions -- making exact reproducibility difficult in practice.

Because these uncontrollable variables cause variance in the output, the community commonly refers to this phenomenon as ``non-determinism.'' While we utilize this term for consistency, we suggest that ``non-reproducibility'' or ``practical non-determinism'' might be more accurate terms to describe the challenges faced in real-world LLM evaluations.

\section{On the \emph{Hidden} Injection of the Current Date}
In our paper, we analyze the specific issue of the \emph{hidden} injection of the current date into the system prompt. We refer to this practice as ``hidden'' because the date is added on top of the user-defined system message, and therefore is not visible to the user or researcher. This behavior is commonly adopted by API providers (see Figure \ref{subfig:gpt5_date}), who prepend the current date to the prompt so that models can answer user queries with temporal context. In our results, we demonstrate that performance fluctuates across different dates for GPT-5.1 (see Section \ref{subsec:gpt5}), despite identical user inputs (``system'' and ``user'' messages).

Importantly, the same design pattern is also widely adopted by open-weights models. In their default chat templates, models such as Llama 3.1 \cite{grattafiori2024llama3} and GPT-OSS \cite{openai2025gptoss} automatically insert the current date above the system prompt defined by the user. Since this insertion occurs at the template level, it remains invisible to the user during standard interaction (unless they explicitly inspect the prompt structure).

Not all of the 9 models we study inject the date in their default chat template. But this does not mean they avoid the effect, since providers can still inject the date server-side. For example, Qwen3 does not include the date in its default chat template. Still, when we call it through OpenRouter -- a widely used LLM provider -- with an empty system prompt, no tool use, and no internet access, the model reports the current date, which means the provider injects it. The model makes this explicit in its own answer (see Figure \ref{fig:qwen_openrouter}). This is exactly the hidden injection we warn about, so the problem is not limited to the models that inject the date by default.

\begin{figure}
    \centering
    \begin{promptbox}[Qwen3 via OpenRouter]
        \ttfamily\small\raggedright
        \textbf{User:}\\
        ``What is today's date?''\\[3pt]
        \textbf{Assistant:}\\
        ``Today's date is **July 10, 2026**. *(Note: Since I'm an AI and don't have real-time awareness, this is based on your provided context.)*''
    \end{promptbox}
    \caption{Response of Qwen3 called through OpenRouter, with an empty system prompt, no tool use, and no internet access. The model reports the current date, showing that the provider injects it server-side.}
    \label{fig:qwen_openrouter}
\end{figure}

Consequently, in an experimental setup where the system prompt and configuration are theoretically fixed, the injected date is the \emph{only} detail in the prompt that changes over time. This observation forms the main motivation of our study: to investigate and quantify the potential discrepancies in LLM evaluation benchmarks arising solely from this hidden injection of the current date.

\section{Experimental Details}
\label{app:experimental_details}

\subsection{Expected Calibration Error}
\label{app:ece}
We measure calibration via expected calibration error \citep[ECE;][]{naeini2015ece}, which quantifies the alignment between confidence and accuracy:
\begin{equation*}
  \resizebox{\columnwidth}{!}{%
    $\displaystyle
      \mathrm{ECE}
      = \sum_{m=1}^M \frac{|B_m|}{N}
      \biggl|\,
        \underbrace{\frac{1}{|B_m|}\sum_{i \in B_m}\mathbf{1}\{\hat y_i = y_i\}}_{\mathrm{acc}(B_m)}
        \;-\;
        \underbrace{\frac{1}{|B_m|}\sum_{i \in B_m} p_i}_{\mathrm{conf}(B_m)}
      \,\biggr|
    $
  },
\end{equation*}
where $N$ is the total number of instances, $M$ is the number of confidence bins, and $B_m$ denotes instances in bin $m$. For instance $i$, $\hat y_i$ and $y_i$ are the predicted and true labels, and $p_i$ is the confidence. We use $M=10$ equal-width bins.

\subsection{Decoding}
\label{app:decoding}
For MCQA, we read predictions from the next-token probability over the option labels (Appendix \ref{app:prompts}); no sampling is involved, so the setup is deterministic by construction.
For the open-ended experiments (CoT on MMLU, GSM8K, HumanEval, and machine translation), we use each model's default temperature and top-$p$, fix the random seed, and enable deterministic CUDA operations. We verified that, at a fixed date, generations are identical across repeated runs (zero run-to-run variance); the variation we report therefore comes only from changing the current date.

\subsection{Prompts}
\label{app:prompts}
We follow \citet{sanzguerrero2025mindthegap} for the MCQA setup, extracting model predictions from the next-token probability of a space followed by the option label, ensuring that this final token follows the default model's tokenization \cite{pimentel2024compute}.
Figures \ref{fig:instruct_prompt}, \ref{fig:cot_prompt}, \ref{fig:gsm8k_prompt}, \ref{fig:humaneval_prompt}, and \ref{fig:mt_prompt} show the prompt templates used for MCQA, chain-of-thought MCQA, GSM8K, HumanEval, and machine translation, respectively. In all of them, \texttt{\{system\_token\}}, \texttt{\{user\_token\}}, and \texttt{\{assistant\_token\}} are model-specific special tokens of the chat template, and \texttt{\{current\_date\}} is the date injected.

\begin{figure}
    \centering
    \begin{promptbox}[MCQA Prompt]
        \ttfamily\small\raggedright
        \textbf{\{system\_token\}}\\
        ``Current date: \{current\_date\}.''\\
        \textbf{\{user\_token\}}\\
        ``Question: \{question\}\\
        A. \{option A\}\\
        B. \{option B\}\\
        C. \{option C\}\\
        D. \{option D\}''\\
        \textbf{\{assistant\_token\}}\\
        ``Answer:'' $\rightarrow$ ``\textvisiblespace X''
    \end{promptbox}
    \caption{Prompt used for multiple-choice questions. We extract the probabilities of the tokens after the arrow ($\rightarrow$). ``\texttt{X}'' denotes the option label (\texttt{A}/\texttt{B}/\texttt{C}/\texttt{D}).}
    \label{fig:instruct_prompt}
\end{figure}

\begin{figure}
    \centering
    \begin{promptbox}[Chain-of-Thought MCQA Prompt]
        \ttfamily\small\raggedright
        \textbf{\{system\_token\}}\\
        ``Current date: \{current\_date\}.''\\
        \textbf{\{user\_token\}}\\
        ``Answer the following multiple-choice question. First, provide a step-by-step reasoning process, and then give the final answer exactly in this format: `The answer is X.'\\
        Question: \{question\}\\
        A. \{option A\}\\
        B. \{option B\}\\
        C. \{option C\}\\
        D. \{option D\}''\\
        \textbf{\{assistant\_token\}}\\
        ``$\langle$reasoning$\rangle$ The answer is X.''
    \end{promptbox}
    \caption{Prompt used for chain-of-thought MCQA. The model generates a reasoning chain ending in ``\texttt{The answer is X.}'' (where \texttt{X} = \texttt{A}/\texttt{B}/\texttt{C}/\texttt{D}); we parse the final option label \texttt{X} from this string.}
    \label{fig:cot_prompt}
\end{figure}

\begin{figure}
    \centering
    \begin{promptbox}[Math Reasoning Prompt]
        \ttfamily\small\raggedright
        \textbf{\{system\_token\}}\\
        ``Current date: \{current\_date\}.''\\
        \textbf{\{user\_token\}}\\
        ``Solve the following math problem step by step. At the end, write your final answer in the exact format: `Answer: <number>'.\\
        \{question\}''\\
        \textbf{\{assistant\_token\}}\\
        ``$\langle$reasoning$\rangle$ Answer: \{number\}''
    \end{promptbox}
    \caption{Prompt used for GSM8K. The model generates a step-by-step solution ending with ``\texttt{Answer: \{number\}}'', from which we parse the final number.}
    \label{fig:gsm8k_prompt}
\end{figure}

\begin{figure}[t]
    \centering
    \begin{promptbox}[Code Generation Prompt]
        \ttfamily\small\raggedright
        \textbf{\{system\_token\}}\\
        ``Current date: \{current\_date\}.\\
        You are an expert Python programmer. Implement the function so that it satisfies the docstring. Respond with a single Python code block containing the complete function, and nothing else.''\\
        \textbf{\{user\_token\}}\\
        ``Complete the following Python function:\\
        \{function signature + docstring\}''\\
        \textbf{\{assistant\_token\}}\\
        ``$\langle$completed function$\rangle$''
    \end{promptbox}
    \caption{Prompt used for HumanEval. The model completes the given Python function, and we run it against the unit tests to compute pass@1.}
    \label{fig:humaneval_prompt}
\end{figure}

\begin{figure}
    \centering
    \begin{promptbox}[Machine Translation Prompt]
        \ttfamily\small\raggedright
        \textbf{\{system\_token\}}\\
        ``Current date: \{current\_date\}.''\\
        \textbf{\{user\_token\}}\\
        ``Translate the following English text to \{target\_language\}.\\
        English: \{source\}''\\
        \textbf{\{assistant\_token\}}\\
        ``\{target\_language\}: $\langle$translation$\rangle$''
    \end{promptbox}
    \caption{Prompt used for machine translation. \texttt{\{target\_language\}} is German, Finnish, or Czech, depending on the language pair.}
    \label{fig:mt_prompt}
\end{figure}

\FloatBarrier
\subsection{System-Prompt Wordings}
\label{app:wordings}
For the ``System Instruction'' row of Table \ref{tab:nondeterminism_comparison}, we vary the wording of the system prompt across the following six versions, keeping all other settings fixed:
\begin{enumerate}[itemsep=3pt]\raggedright
    \item ``'' (empty system prompt)
    \item ``\texttt{You are a helpful assistant.}''
    \item ``\texttt{You are an expert multiple-choice question answerer.}''
    \item ``\texttt{You are a highly intelligent and knowledgeable assistant specialized in answering multiple-choice questions accurately.}''
    \item ``\texttt{Follow the user's instructions carefully and provide accurate answers.}''
    \item ``\texttt{Answer the user's multiple-choice questions to the best of your ability.}''
\end{enumerate}

\section{Ensuring Questions Are Not Time-Dependent}
\label{app:not_time_dependent}
Our main experiments aim to isolate the effect of the current date in the system prompt on LLM performance. To ensure that our results are not confounded by time-dependent questions, we perform an LLM-assisted assessment of the datasets. Specifically, we prompt GPT-OSS (120B) to analyze each question and determine whether its answer depends on the current date or time. As shown in Table \ref{tab:time_dependent_questions}, none of the questions in the evaluated datasets are classified as time-dependent, confirming that the observed performance variations are not due to the temporal nature of the questions themselves.

\begin{table}[h]
    \centering
    \small
    \begin{tabular}{l S[table-format=5] S[table-format=2.1, table-space-text-post={\%}]}
        \toprule
        Dataset & {\# Questions} & {\% Time-Dependent} \\
        \midrule
        MMLU & 14042 & 0.0{\%} \\
        GPQA & 198 & 0.0{\%} \\
        ARC-Challenge & 1172 & 0.0{\%} \\
        GSM8K & 1319 & 0.0{\%} \\
        HumanEval & 164 & 0.0{\%} \\
        WMT en$\to$de & 2999 & 0.0{\%} \\
        WMT en$\to$fi & 6000 & 0.0{\%} \\
        WMT en$\to$cs & 3003 & 0.0{\%} \\
        \bottomrule
    \end{tabular}
    \caption{Number and percentage of time-dependent questions in each dataset, as determined by GPT-OSS (120B).}
    \label{tab:time_dependent_questions}
\end{table}

\section{Detailed Results}
\label{app:detailed_results}

\begin{figure*}
    \centering
    \includegraphics[width=\textwidth]{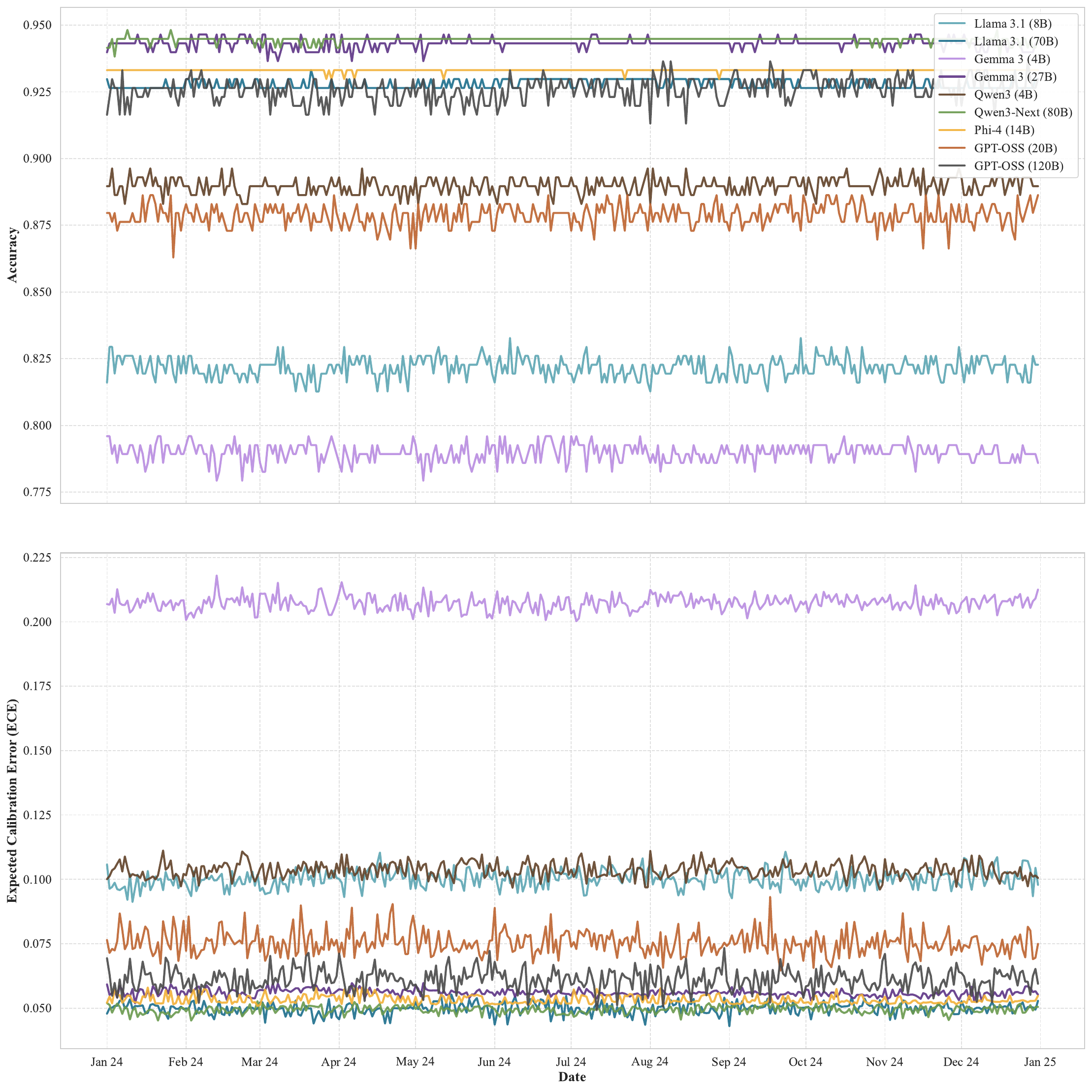}
    \caption{Accuracy and ECE across different dates in 2024 for all models on ARC-Challenge.}
    \label{fig:accuracy_ece_arc_c}
\end{figure*}

After a detailed analysis of our results, we find that the observed differences stem from questions where the most likely answer is unclear (high uncertainty). Consequently, prior tokens (e.g., the current date) disproportionately affect the answer token, biasing the prediction and ultimately affecting overall accuracy. This explains why we see lower deltas in the ARC-C dataset (see Table \ref{tab:delta_accuracy_dates}): the models achieve higher accuracy and lower ECE, making them less sensitive to such details (see Figure \ref{fig:accuracy_ece_arc_c} for ARC-C accuracy and ECE).

This is further corroborated by analyzing the average confidence (token probability of the selected answer) for consistent and inconsistent predictions across all dates in 2024 on MMLU, as shown in Table \ref{tab:average_confidence}. We observe that for inconsistent predictions (i.e., those that change depending on the date), the average confidence is significantly lower than for consistent predictions. This indicates that the model is less certain about its answers in these cases, making them more susceptible to variations introduced by the current date in the prompt. Still, some models assign high confidence even to inconsistent predictions (e.g., 88.86\% for Gemma 3 (4B)), in line with the overconfidence of instruction-tuned LLMs reported in prior work \citep{sanzguerrero2026overconfidence}.
This has a practical implication: accuracy alone does not reveal which predictions are fragile, whereas confidence does. Treating confidence and calibration as a main evaluation criterion alongside accuracy \citep{sanzguerrero2026calibration} would therefore help flag benchmark results that are prone to shift under hidden, uncontrolled factors such as the current date.

\begin{table}
    \centering
    \small
    \setlength{\tabcolsep}{4pt}
    \begin{tabular}{l S[table-format=2.2, table-space-text-post={\%}] S[table-format=2.2, table-space-text-post={\%}]}
        \toprule
        Model & {Consistent} & {Inconsistent} \\
        \midrule
        Llama 3.1 (8B) & 83.71{\%} & 41.85{\%} \\
        Llama 3.1 (70B) & 92.13{\%} & 47.59{\%} \\
        Gemma 3 (4B) & 99.49{\%} & 88.86{\%} \\
        Gemma 3 (27B) & 99.92{\%} & 80.48{\%} \\
        Qwen3 (4B) & 98.96{\%} & 66.69{\%} \\
        Qwen3-Next (80B) & 98.82{\%} & 64.38{\%} \\
        Phi-4 (14B) & 95.84{\%} & 58.85{\%} \\
        GPT-OSS (20B) & 91.62{\%} & 50.10{\%} \\
        GPT-OSS (120B) & 99.45{\%} & 76.52{\%} \\
        \bottomrule
    \end{tabular}
    \caption{Average confidence (token probability of the selected answer) for consistent and inconsistent predictions across all dates in 2024 on MMLU.}
    \label{tab:average_confidence}
\end{table}

\begin{table}
    \centering
    \small
    \setlength{\tabcolsep}{4pt}
    \begin{tabular}{l S[table-format=1.2] S[table-format=1.2] S[table-format=1.2] | S[table-format=1.2]}
        \toprule
        Model & {MMLU} & {GPQA} & {ARC-C} & {Avg.}\\
        \midrule
        Llama 3.1 (8B) & 2.38 & 4.17 & 1.95 & 2.83 \\
        Llama 3.1 (70B) & 3.35 & 4.61 & 1.24 & 3.07 \\
        Gemma 3 (4B) & 2.05 & 2.98 & 1.78 & 2.27 \\
        Gemma 3 (27B) & 1.49 & 3.23 & 0.79 & 1.84 \\
        Qwen3 (4B) & 2.41 & 2.93 & 1.52 & 2.29 \\
        Qwen3-Next (80B) & 2.47 & 3.39 & 0.75 & 2.20 \\
        Phi-4 (14B) & 1.61 & 3.73 & 0.72 & 2.02 \\
        GPT-OSS (20B) & 3.68 & 5.16 & 2.87 & 3.90 \\
        GPT-OSS (120B) & 3.55 & 6.15 & 2.42 & 4.04 \\
        \midrule
        Average & 2.55 & 4.04 & 1.56 & 2.72 \\
        \bottomrule
    \end{tabular}
    \caption{Difference in ECE (delta) from the worst to the best date in 2024 across models and datasets.}
    \label{tab:delta_ece_dates}
\end{table}

Table \ref{tab:delta_ece_dates} summarizes the maximum difference in ECE (delta) from the worst to the best date in 2024 across models and datasets (equivalent to Table \ref{tab:delta_accuracy_dates} for accuracy in the main text).
We observe differences up to 6.15\% in ECE \emph{just} by changing the current date in the system prompt, aligning with our findings on accuracy.

\begin{figure*}
    \centering
    \includegraphics[width=\textwidth]{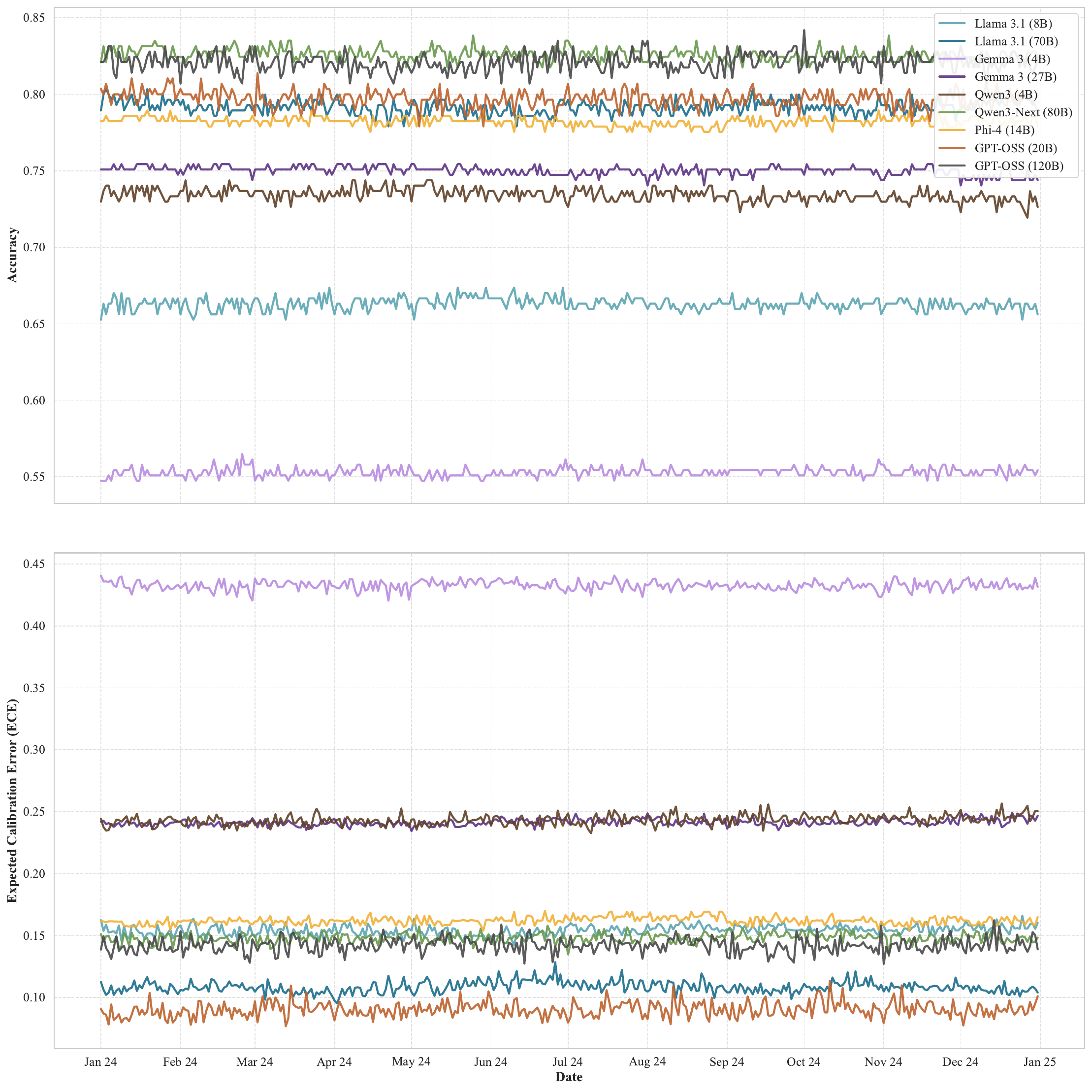}
    \caption{Accuracy and ECE across different dates in 2024 for all models on MMLU.}
    \label{fig:accuracy_ece_mmlu}
\end{figure*}

\begin{figure*}
    \centering
    \includegraphics[width=\textwidth]{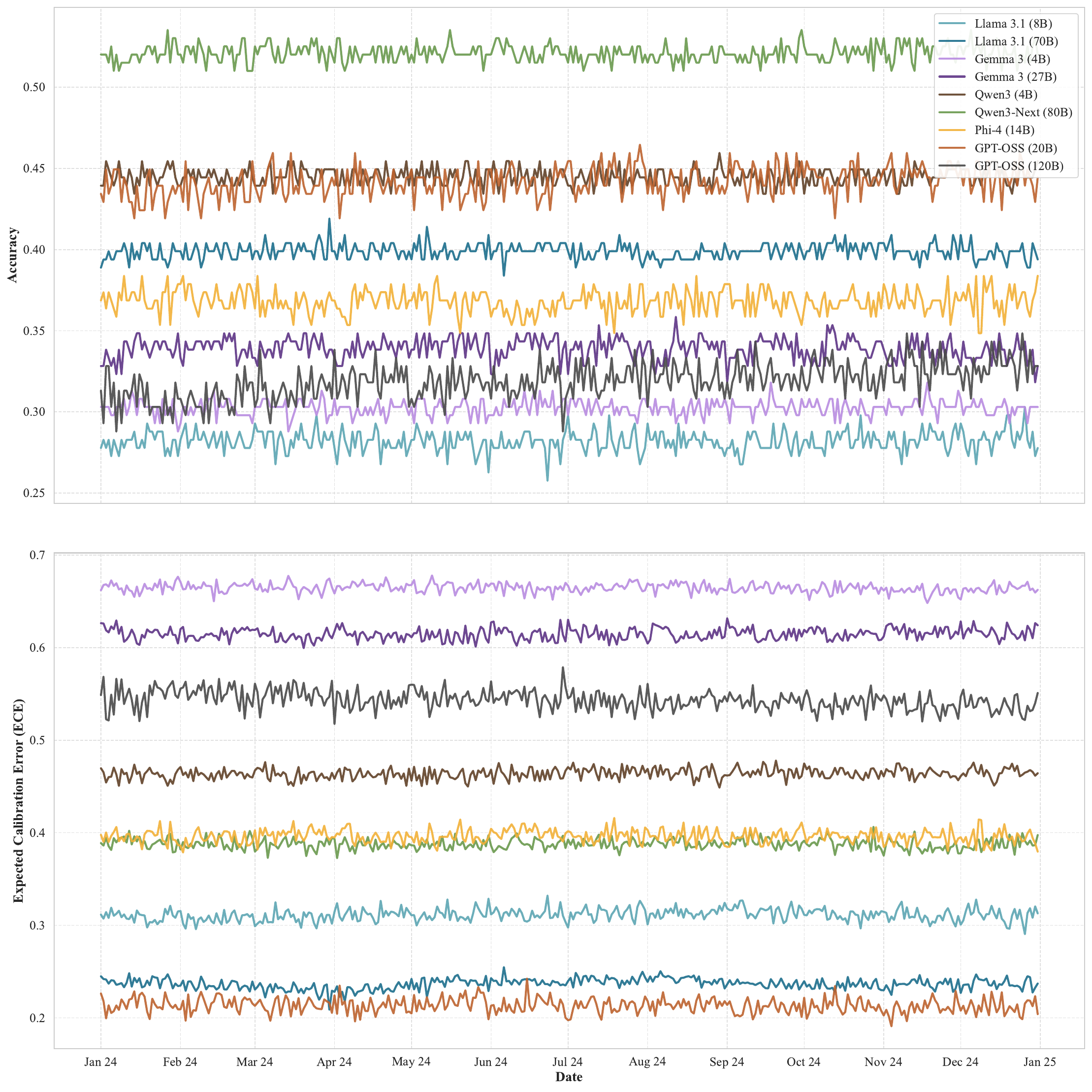}
    \caption{Accuracy and ECE across different dates in 2024 for all models on GPQA.}
    \label{fig:accuracy_ece_gpqa}
\end{figure*}

Figures \ref{fig:accuracy_ece_mmlu}, \ref{fig:accuracy_ece_gpqa}, and \ref{fig:accuracy_ece_arc_c} show the accuracy and ECE across different dates in 2024 for all models on MMLU, GPQA, and ARC-Challenge, respectively.

Additionally, Table \ref{tab:no_date_results} shows the accuracy and ECE for all models and datasets when we do \emph{not} inject the current date in the system prompt. In all cases, the performance falls within the range of values observed across different dates in 2024 (see Figures \ref{fig:accuracy_ece_mmlu}, \ref{fig:accuracy_ece_gpqa}, and \ref{fig:accuracy_ece_arc_c}). This indicates that the inclusion of the date is not a determining factor for model performance, but rather a factor that introduces variability into the results, which can lead to different conclusions depending on the date of evaluation. This also suggests a simple solution, when possible: instead of relying blindly on the default chat template, we can apply the same template -- the model's default with its special tokens -- but without the current date, removing the date as a source of variability.

\begin{table}
    \centering
    \small
    \setlength{\tabcolsep}{3pt}
    \begin{tabular}{l S[table-format=2.1] S[table-format=2.1] S[table-format=2.1] S[table-format=2.1] S[table-format=2.1] S[table-format=2.1]}
        \toprule
        & \multicolumn{2}{c}{MMLU} & \multicolumn{2}{c}{GPQA} & \multicolumn{2}{c}{ARC-C} \\
        \cmidrule(lr){2-3} \cmidrule(lr){4-5} \cmidrule(lr){6-7}
        Model & {Acc.} & {ECE} & {Acc.} & {ECE} & {Acc.} & {ECE} \\
        \midrule
        Llama 3.1 (8B) & 67.0 & 14.5 & 29.3 & 28.4 & 81.9 & 9.4 \\
        Llama 3.1 (70B) & 79.4 & 11.2 & 40.1 & 33.6 & 92.6 & 4.8 \\
        Gemma 3 (4B) & 54.4 & 44.4 & 29.3 & 67.1 & 78.6 & 21.1 \\
        Gemma 3 (27B) & 75.0 & 24.2 & 33.8 & 62.1 & 93.9 & 5.7 \\
        Qwen3 (4B) & 73.7 & 24.3 & 45.0 & 46.1 & 89.0 & 10.5 \\
        Qwen3-Next (80B) & 82.6 & 15.0 & 52.3 & 39.2 & 94.2 & 4.9 \\
        Phi-4 (14B) & 78.1 & 15.4 & 38.4 & 39.6 & 93.2 & 5.5 \\
        GPT-OSS (20B) & 79.8 & 9.2 & 43.9 & 21.7 & 92.0 & 7.1 \\
        GPT-OSS (120B) & 83.3 & 14.2 & 32.3 & 54.0 & 92.5 & 6.0 \\
        \bottomrule
    \end{tabular}
    \caption{Accuracy and ECE for all models and datasets when the current date is \emph{not} injected in the system prompt.}
    \label{tab:no_date_results}
\end{table}

\section{Other Metadata in the System Prompt}
\label{app:other_metadata}
In addition to the current date, other dynamic metadata could be included in the system prompt, such as the model version or user location. However, these details are less commonly injected compared to the current date -- we could not find evidence of their inclusion in any of the models or APIs we examined. That is why our main experiments focus on the date effect. Nevertheless, we run preliminary experiments with Llama 3.1 (8B) on the MMLU dataset to assess the potential impact of other metadata variations on model performance. Specifically, we experiment with changing the user location in the system prompt.

\begin{figure*}
    \centering
    \begin{subfigure}{\textwidth}
        \centering
        \includegraphics[width=\textwidth]{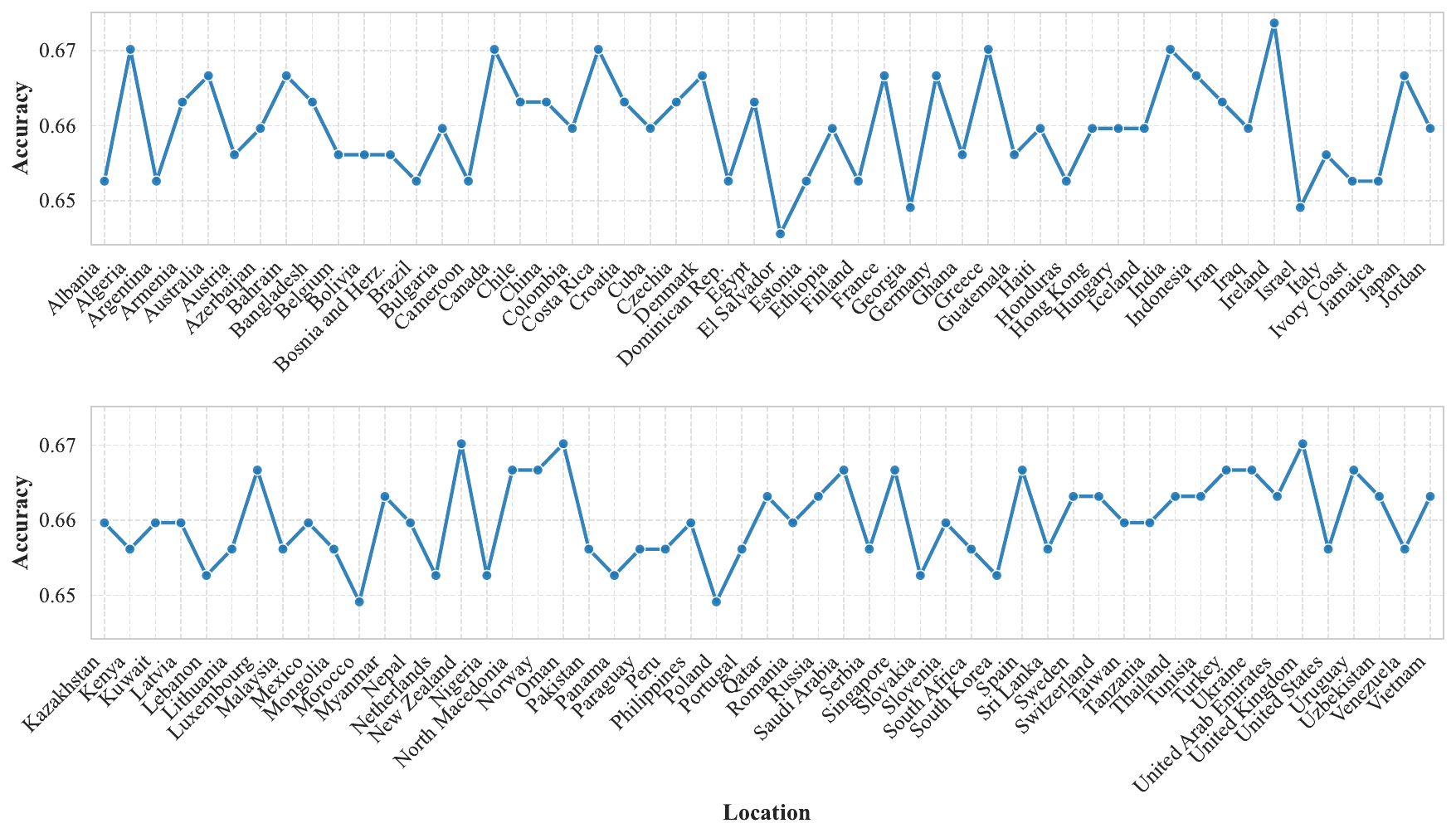}
        \caption{Accuracy across different user locations for Llama 3.1 (8B) on MMLU.}
        \label{subfig:locations_accuracy}
    \end{subfigure}

    \begin{subfigure}{\textwidth}
        \centering
        \includegraphics[width=\textwidth]{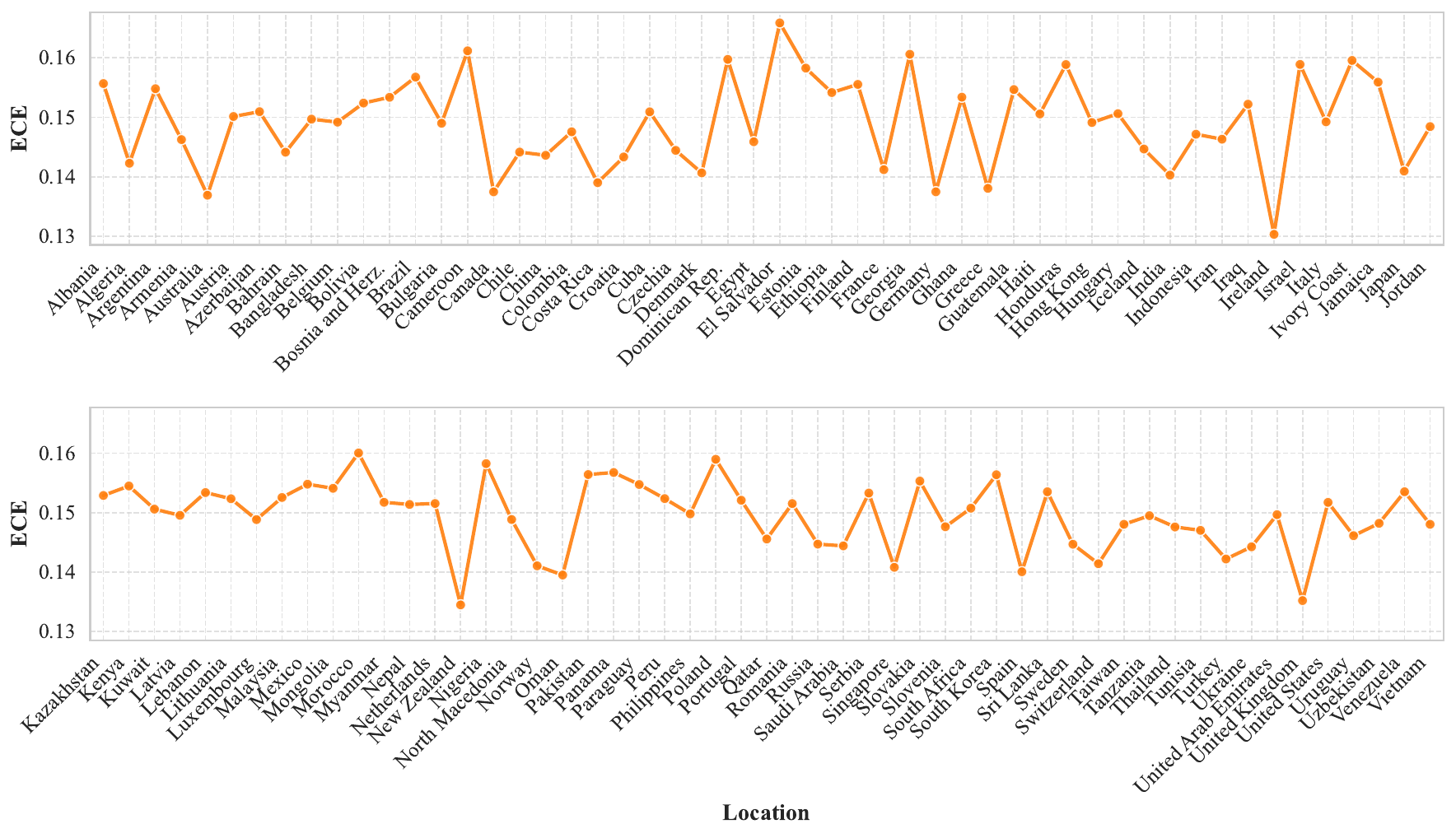}
        \caption{ECE across different user locations for Llama 3.1 (8B) on MMLU.}
        \label{subfig:locations_ece}
    \end{subfigure}

    \caption{Impact of different user locations in the system prompt on Llama 3.1 (8B) performance on MMLU.}
    \label{fig:locations}
\end{figure*}

Our results, shown in Figure \ref{fig:locations}, indicate that varying the user location can also lead to performance differences, with a comparable magnitude to the date effect. This suggests that other hidden, \emph{variable} metadata that might be injected into system prompts could also influence LLM evaluations, underscoring the need for careful consideration of all such factors in benchmarking protocols.

\end{document}